\documentclass{article}

\usepackage{arxiv}

\usepackage[utf8]{inputenc} 
\usepackage[T1]{fontenc}    
\usepackage{hyperref}       
\usepackage{url}            
\usepackage{booktabs}       
\usepackage{amsfonts}       
\usepackage{nicefrac}       
\usepackage{microtype}      
\usepackage{graphicx}
\usepackage{subfigure}

\title{Instance Segmentation Challenge Track Technical Report, VIPriors Workshop at ICCV 2021: Task-Specific Copy-Paste Data Augmentation Method for Instance Segmentation}

\author{ Jahongir Yunusov \\
	DIVUS corp., Republic of Korea\\
	\texttt{jahongir7174@gmail.com} \\
	\And
    Shohruh Rakhmatov \\
	DeltaX corp., Republic of Korea\\
	\texttt{shoh.mirzo.786@gmail.com} \\
	\And
    Abdulaziz Namozov \\
	DP World, Republic of Korea\\
	\texttt{anamozov@gmail.com} \\
	\And
    Abdulaziz Gaybulayev \qquad \qquad Tae-Hyong Kim\\
	Department of Computer Engineering \\
	Kumoh National Institute of Technology, Republic of Korea\\
	\texttt{\{g.abdulaziz, taehyong\}@kumoh.ac.kr} \\
}

\date{}

\hypersetup{
pdftitle={Task-Specific Copy-Paste Data Augmentation Method for Instance Segmentation},
pdfkeywords={VIPriors Workshop Challenge, Instance Segmentation, Copy-Paste Augmentation},
}

\begin{document}
\maketitle

\begin{abstract}
Copy-Paste has proven to be a very effective data augmentation for instance segmentation which can improve the generalization of the model \cite{Ghiasi2021}. We used a task-specific Copy-Paste data augmentation method to achieve good performance on the instance segmentation track of the 2nd VIPriors workshop challenge. We also applied additional data augmentation techniques including RandAugment and GridMask. 
Our segmentation model is the HTC detector on the CBSwin-B with CBFPN with some tweaks. This model was trained at the multi-scale mode by a random sampler on the 6x schedule and tested at the single-scale mode.  By combining these techniques, we achieved 0.398 AP@0.50:0.95 with the validation set and 0.433 AP@0.50:0.95 with the test set. Finally, we reached 0.477 AP@0.50:0.95 with the test set by adding the validation set to the training data. Source code is available at https://github.com/jahongir7174/VIP2021.
\end{abstract}


\section{Introduction}

Instance segmentation \cite{Dai2016} is a useful computer vision task widely used in many real-world applications such as autonomous driving and medical diagnostics. Like other deep learning-based tasks, deep learning-based instance segmentation requires a large training dataset to achieve good performance. Moreover, annotating a large dataset for instance segmentation is more time-consuming than those of other tasks. To address these deep learning issues, the 2nd visual inductive priors for data-efficient deep learning workshop is held at ICCV 2021. Additionally, the workshop offers five computer vision task challenges, including instance segmentation, where models are to be trained from scratch in a data-deficient setting. It is prohibited to use other datasets than the provided training data, so no pre-training and no transfer learning is allowed in every track of the competition. 

In the instance segmentation challenge, Synergy Sports, challenge co-organizer, provided a dataset consisting of 184, 62, and 64 images in train, validation, and test sets, respectively. The images are from basketball game scenes, where polygon-based annotations are provided for players and basketball only. 
In order to achieve satisfactory segmentation accuracy using that small dataset, we tested various state-of-the-art segmentation network models and data augmentation techniques.


The rest of the paper is organized as follows. Section 2 briefly introduces the related work, and section 3 describes the proposed data augmentation method. Then, experiment results are presented in section 4.

\section{Related Work}
\label{sec:realtedwork}


{\bf Data Augmentation.} General-purpose data augmentation techniques including color-space \cite{Szegedy2015} and geometric \cite{Cubuk2020} transformations are widely used and have shown great classification performance improvements with the ImageNet dataset\cite{Russakovsky2015}. Recently, special data transformation methods have been proposed as data augmentation methods, such as Mixup \cite{He2017} in which two corresponding images are mixed at an arbitrary ratio. Of those methods, Copy-paste \cite{Zhang2018}, which copies object from one image to another, is particularly useful for instance segmentation.

{\bf Backbone Network.} When constructing deep learning networks for computer vision tasks, backbone models play an important role in determining performance by capturing features from images. Numerous CNN-based backbones were proposed for classification, including AlexNet\cite{Krizhevsky2012}, ResNet\cite{He2016}, ResNeXt\cite{Xie2017} and Res2Net\cite{Gao2021}. Recently Transformer models, which demonstrated remarkable performances in natural language processing, have been applied to computer vision tasks and have shown succussful results especially in object detection such as Transformer \cite{Vaswani2017}, CNN-Transformer DETR \cite{Carion2020} and Swin Transformer \cite{Liu2021}. CBNetV2 \cite{Liang2021}, the current latest model, combines high- and low-level features for getting better efficient object detector. This model achieved state-of-the-art performance in both object detection and instance segmentation on the COCO \cite{Lin2014} benchmark dataset.

{\bf Instance Segmentation.} Instance segmentation \cite{Liang2021,He2016} detects all instances of the target objects and splits the pixels of each detected instance. Hybrid Task Cascade \cite{Ghiase2021} is a recent state-of-the-art model which has a special cascade architecture for instance segmentation.

\section{Data Augmentation}
\label{sec:method}

\begin{figure}%
\centering

\mbox{%
	\subfigure[][Original input image]{%
	\includegraphics[width=0.48\textwidth]{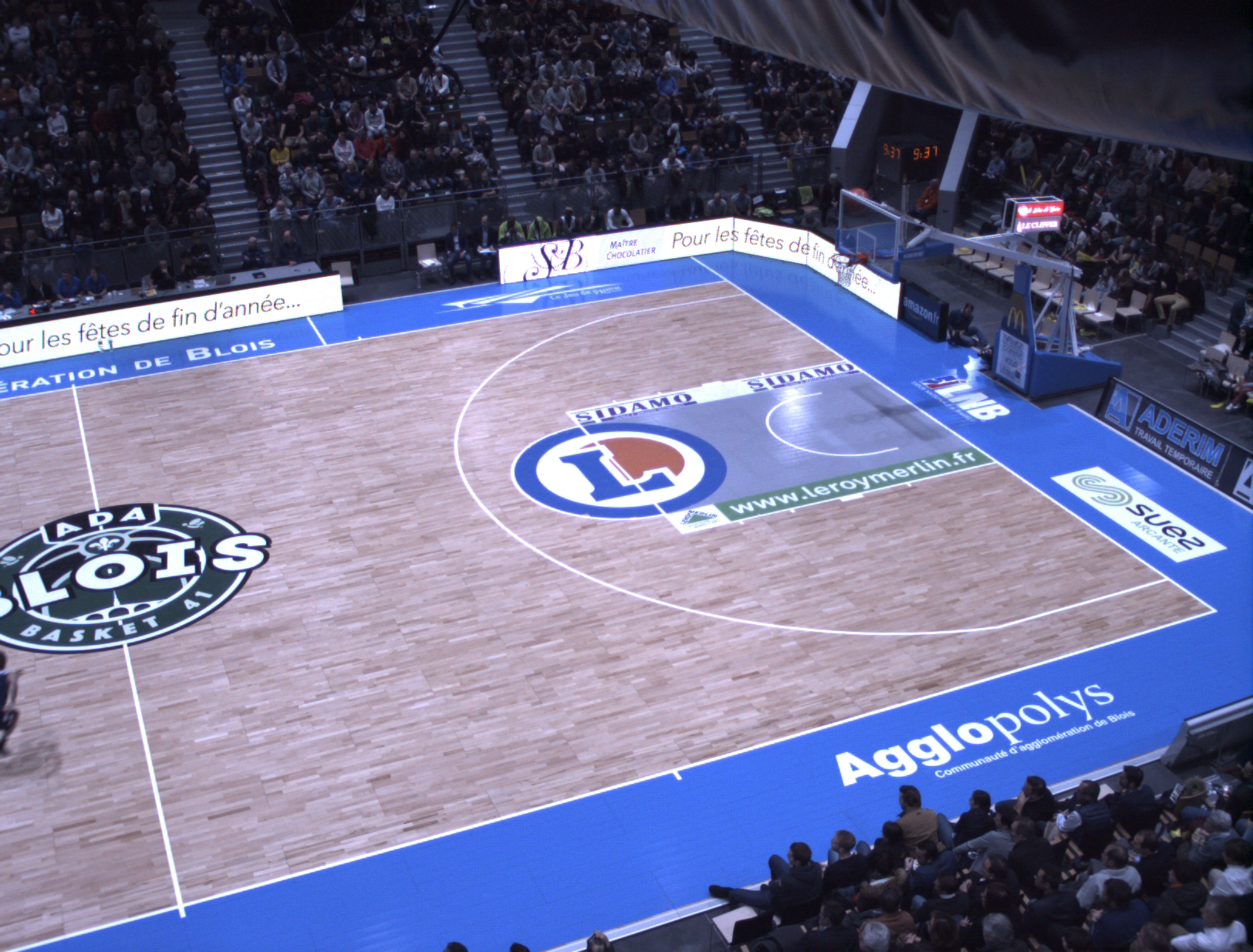}}%
	\quad

	\subfigure[][Copy-paste data augmentation]{%
	\includegraphics[width=0.48\textwidth]{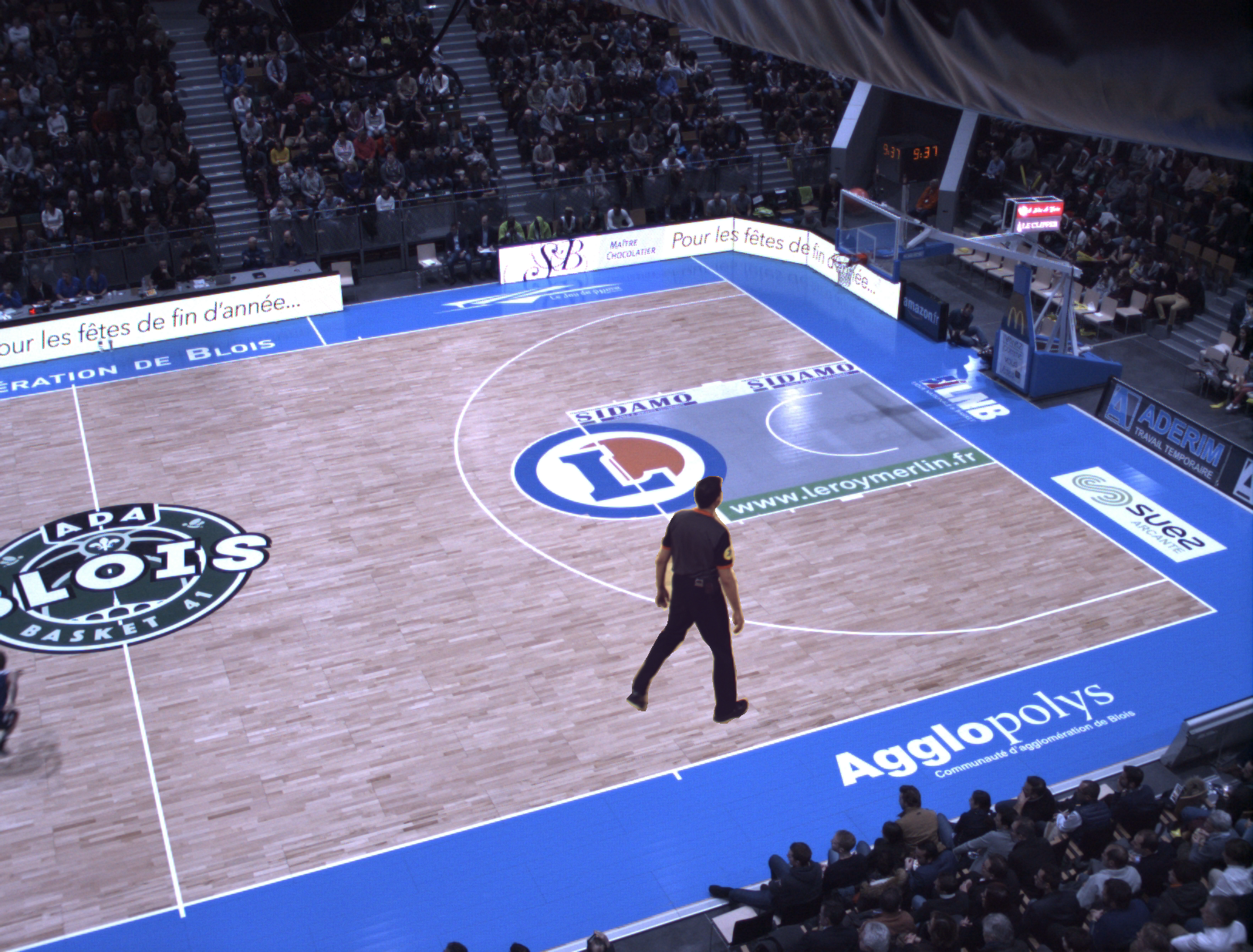}}%
}
\mbox{%
	\subfigure[][Grid-mask data augmentation]{%
	\includegraphics[width=0.48\textwidth]{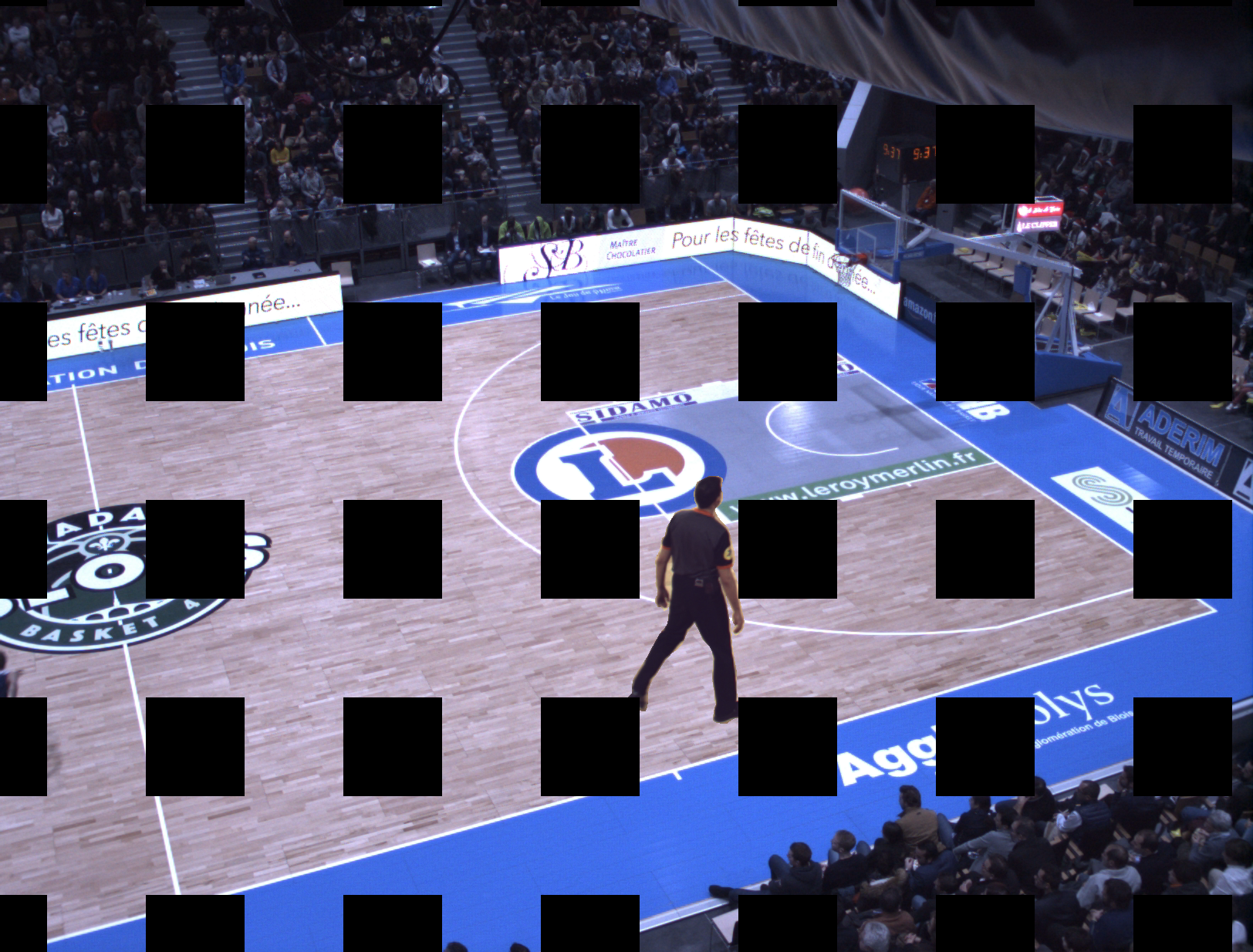}}%
	\quad

	\subfigure[][Random data augmentation]{%
	\includegraphics[width=0.48\textwidth]{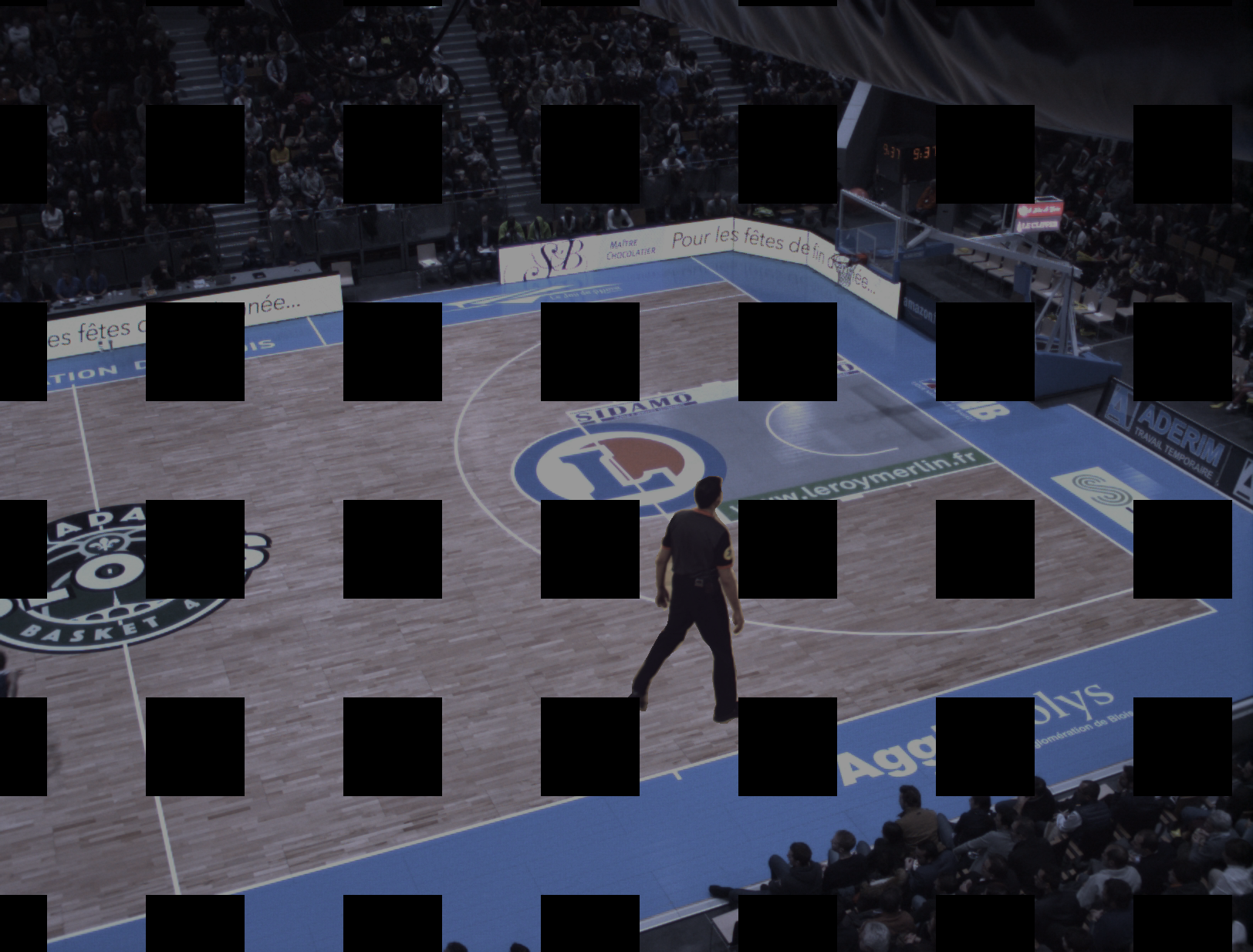}}%
}
	\caption{Visualization of three data augmentation methods.}
	\label{fig:vis}
\end{figure}

Although our data augmentation approach is simple, it is effective in improving the segmentation results and generalizing the instance segmentation model on the given dataset. First, we cropped all objects, then saved the cropped images and their corresponding mask annotations to storage. Then we duplicated images 20 times without any change. All duplicated images are used as input to our copy-paste augmentation method during training. In order to maintain class balance, all class instances are randomly sampled from 5 to 15. Unlike the original copy-paste data augmentation \cite{Zhang2018}, we paste each object, a player or a ball, into a specific location where it should be. The random copy-paste location constraint is as follows:

\begin{equation}
	256 \leq x_{min} \leq w-256
\end{equation}

\begin{equation}
	\frac{w}{2} + 256 \leq y_{min} \leq \frac{h}{2} + 256
\end{equation}

, where $w$ and $h$ are the width and the height of an image, respectively, and ($x_{min}$, $y_{min}$) denotes the top left coordinate of an object that is being pasted.
Figure \ref{fig:vis} shows an example of our data augmentation method.

\section{Experiments}
\label{sec:experiments}

In this section, we show the results of some experiments on the provided instance segmentation dataset and compare different methods to verify effectiveness of our method. 

The baseline model is the HTC detector \cite{Chen2019} on the CBSwin-T backbone with CBFPN \cite{Liang2021} using group normalization, trained at the multi-scale mode by a random sampler on the 2x schedule \cite{Chen2019b}. Note that all ReLU \cite{Nair2010} activation functions used in the box and mask heads have been replaced with sigmoid linear unit (SiLU) \cite{Ramachandran2017} functions. 

In training the original model, the input image is randomly scaled from 400 to 1400 on the short side and up to 1600 on the long side. We adjusted the minimum size of the short side to 800. In our test phase, the single scale (1600, 1400) was used without any test time augmentation (TTA).  
Except for the last submission, experiments were run on two NVIDIA V100 GPUs. Our last submission used three NVIDIA V100 GPUs.

The detailed step-by-step results of our experiments on validation and test set is described in Table \ref{tab:1}. 
In Table \ref{tab:1}, TS Copy-Paste refers to the task-specific copy-paste data augmentation technique described in section \ref{sec:method}. In Better backbone setup, we replaced CBSwin-T with CBSwin-B and applied 6x schedule \cite{Chen2019b}. With the three augmentation methods and the replaced backbone, we achieved 0.398 AP@0.50:0.95 with the validation set and 0.433 AP@0.50:0.95 with the test set. A better test AP might show that the test set has a more similar distribution to the train set than the validation set. Finally, we reached 0.477 AP@0.50:0.95 with the test set by adding the validation set to the training data.

\begin{table}
	\caption{Step-by-step results on validation and test set}
	\centering
	\begin{tabular}{lccc}
		\toprule
		Methods     & Schedule & Val AP@0.50:0.95 & Test AP@0.50:0.95\\
		\midrule
		Baseline  & 2x & 0.186  & -     \\
		+ TS Copy-Paste & 2x  & 0.338  & -      \\
		+ RandAugment \cite{Cubuk2020} + GridMask \cite{Chen2020}  & 2x & 0.345  &  -  \\
		+ Better backbone & 6x & 0.398   & 0.433  \\
		+ Val set added in training  & 6x &  -  & 0.477  \\
		\bottomrule
	\end{tabular}
	\label{tab:1}
\end{table}

\bibliographystyle{unsrtnat}


\end{document}